\documentclass[10pt,twocolumn,letterpaper]{article}
\usepackage[pagenumbers]{cvpr}
\usepackage{amsmath,amssymb,amsthm,graphicx,booktabs,multirow,pifont}
\definecolor{cvprblue}{rgb}{0.21,0.49,0.74}
\usepackage[pagebackref,breaklinks,colorlinks,allcolors=cvprblue]{hyperref}
\newcommand{\E}{\mathbb E}

\newcommand{\DIGN}{\textsc{DIGN}}
\AtBeginEnvironment{thebibliography}{\interlinepenalty=10000\relax}
\def\confName{CVPR}
\def\confYear{2027}
\title{DIGN: Divergent Gradient Response via Noise for Adversarial Robustness}
\author{Zhiyuan Ye, Yueguo Wang, and Yi Zhou\\
University of Science and Technology of China\\
{\small \href{mailto:zhiyuan.ye@mail.ustc.edu.cn}{\nolinkurl{zhiyuan.ye@mail.ustc.edu.cn}}\quad \href{mailto:shiyan688@mail.ustc.edu.cn}{\nolinkurl{shiyan688@mail.ustc.edu.cn}}\quad \href{mailto:yi_zhou@ustc.edu.cn}{\nolinkurl{yi_zhou@ustc.edu.cn}}}}
\hypersetup{
  pdftitle={DIGN: Divergent Gradient Response via Noise for Adversarial Robustness},
  pdfauthor={Zhiyuan Ye, Yueguo Wang, Yi Zhou}
}
\begin{document}
\maketitle
\begin{abstract}
Deep neural networks remain vulnerable to adversarial examples: small input perturbations can change their predictions. Conventional adversarial training seeks correctness throughout a local perturbation set. This objective places stringent demands on the decision boundary, often requiring model capacity and training data substantially beyond the budget needed for high clean accuracy on the same task.

Adding defense components to existing models is a common low-cost way to make attacks harder. However, many defenses mask gradients and fail when adaptive attacks reliably estimate them. We propose DIGN (Divergent Gradient Response via Noise), which adversarially optimizes lightweight pre-activation residual adapters under Gaussian input noise while freezing the robust backbone and batch normalization. Unfiltered diagnostics reveal nearly orthogonal cross-noise gradients. Post-adaptation bilateral filtering strengthens robustness at higher sampling budgets. Evaluations on CIFAR-10, CIFAR-100, and Tiny-ImageNet using PGD, AutoAttack, and other attacks show that DIGN brings adversarial accuracy close to clean accuracy, retaining high robustness under the evaluated adaptive and black-box attacks. Ablations further show that, among the evaluated configurations, the components improve model robustness synergistically only under the proposed design.
\end{abstract}

\section{Introduction}
\begin{figure}[t]
\centering
\includegraphics[width=.98\linewidth]{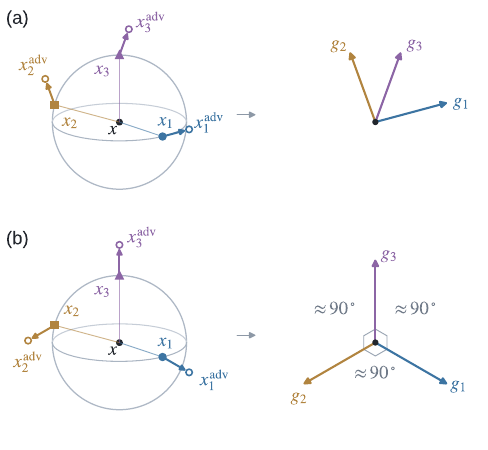}
\caption{Cross-noise gradient responses of noise only (a) and DIGN (b).}
\label{fig:schematic}
\end{figure}

Deep neural networks recognize clean images accurately, yet small, visually inconspicuous input perturbations can mislead them~\cite{szegedy2014intriguing,goodfellow2015explaining}. Adversarial training repeatedly generates adversarial examples and optimizes the classifier to maintain correct predictions within a prescribed perturbation set~\cite{madry2018towards,zhang2019trades}. This objective places greater demands on the decision boundary than clean classification. Larger models and more training data can improve adversarial robustness~\cite{gowal2020uncovering,carmon2019unlabeled,wang2023better}, but scaling also increases data and computational costs. This motivates a complementary question: how can an existing model gain further robustness with few additional trainable parameters?

Input processing and randomization offer a common way to make attacks on existing models harder~\cite{xie2018mitigating,raff2019barrage,nie2022diffpure}. However, lower attack success does not always reflect a more robust classifier: unreliable backward passes or stochastic forward passes can mask gradients and fail under attacks adapted to the defense~\cite{athalye2018obfuscated,tramer2020adaptive}. Given a known randomization distribution, an attacker can estimate the expected-loss gradient through expectation over transformation (EOT)~\cite{athalye2018eot}. A defense must therefore improve predictions under random inputs while remaining effective when the attacker knows the entire pipeline and adapts the attack accordingly. Adding noise alone does not automatically achieve this; the model must learn to use noisy observations reliably.

We propose \DIGN{} (Divergent Gradient Response via Noise), which learns internal responses adapted to input noise in an existing robust model. We freeze all original backbone weights and batch-normalization parameters and statistics, insert lightweight channel-wise residual MLPs before activations, and adversarially optimize them under Gaussian input noise. Each scalar map is shared across spatial positions, introducing neither spatial nor channel mixing; WRN-28-4 adds only approximately 6\% to the original backbone parameter count. Zero-output initialization preserves the model function upon insertion, while subsequent adaptation can adjust activations, decision boundaries, and input derivatives. Adaptation uses no additional training images, and deployment uses a single random prediction without voting.

Randomness alone is insufficient. As illustrated in Fig.~\ref{fig:schematic}(a), adding noise changes the input, yet local attack directions across noisy observations can remain aligned, allowing a shared perturbation to affect multiple observations. Figure~\ref{fig:schematic}(b) illustrates DIGN's key observation: after learning internal adapters, input gradients across noisy observations exhibit nearly orthogonal directional responses. Consistent predictions under noise need not require consistent local gradients. The existing unfiltered gradient diagnostic supports this observation (Fig.~\ref{fig:gradients}), although no explicit gradient-orthogonality loss is used to obtain it.

After adaptation, we add bilateral filtering~\cite{tomasi1998bilateral,stalder2026noise} to strengthen robustness at higher attack sampling budgets. Full-pipeline attacks show that the unfiltered model loses accuracy as EOT samples increase, whereas filtering retains higher accuracy on the same test images over the evaluated budgets (Fig.~\ref{fig:eot}). We thus examine both the gradient responses associated with noise adaptation and the further benefit of filtering. The final system is evaluated with complete stochastic forward passes, exact gradients, and supplementary adaptive and black-box attacks.

Evaluations on CIFAR-10, CIFAR-100, and Tiny-ImageNet using PGD, AutoAttack, and other attacks show that DIGN brings adversarial accuracy close to clean accuracy. Experiments cover different WRN widths and both TRADES and PGD-AT pretraining objectives; high robustness is retained under the evaluated adaptive and black-box attacks. Ablations further show that simply adding noise, training adapters alone, or jointly training from scratch does not match the proposed configuration. The synergy depends on the adaptation design rather than randomness or additional capacity alone.

Our contributions are threefold:
\begin{itemize}
\item \textbf{Parameter-efficient robustness enhancement of existing models.} DIGN combines input randomization with activation adaptation on a frozen backbone. Without retraining the backbone or using additional data, it substantially improves WRN-28-4 adversarial robustness with only approximately 6\% additional parameters.
\item \textbf{Prediction consistency can coexist with gradient divergence.} We find that adversarial adaptation under noise can retain reliable predictions while exhibiting nearly orthogonal cross-noise gradient responses. This behavior is much less pronounced in noise-only and joint-training controls, revealing a substantive difference between learning stochastic responses and merely introducing randomness.
\item \textbf{A substantially narrower clean--robust accuracy gap.} On CIFAR-10, CIFAR-100, and Tiny-ImageNet, DIGN brings adversarial accuracy close to clean accuracy and retains high robustness under the evaluated adaptive and black-box attacks. Ablations further show synergistic gains from noise, adapters, and filtering under the proposed design, outperforming individual components and joint training from scratch.
\end{itemize}

\section{Related Work}
\paragraph{{\spaceskip=\fontdimen2\font\relax Attacks and adversarial training.}}
Adversarial examples expose prediction failures under small input changes~\cite{szegedy2014intriguing,goodfellow2015explaining}. PGD, DeepFool, C\&W, and FAB represent complementary optimization attacks~\cite{madry2018towards,moosavi2016deepfool,carlini2017towards,croce2020fab}. PGD adversarial training and TRADES provide our basic objectives~\cite{madry2018towards,zhang2019trades}. Free/Fast AT reduce training cost~\cite{shafahi2019adversarial,wong2020fast}, while robust-overfitting analysis, AWP, and objective design address training and generalization~\cite{rice2020overfitting,wu2020adversarial,pang2022robustness}. Robust generalization can require more data, while standard and robust accuracy can conflict on some distributions~\cite{schmidt2018data,tsipras2019robustness}. Additional real or generated data provide another route~\cite{carmon2019unlabeled,gowal2020uncovering,rebuffi2021fixing,wang2023better}. DIGN instead adapts existing weights through newly inserted modules, without additional training images.

\paragraph{{\spaceskip=\fontdimen2\font\relax Input processing and randomization.}}
Bilateral filtering is a classical edge-aware operation~\cite{tomasi1998bilateral}. Input randomization, transform ensembles, PixelDefend, and feature denoising modify inputs or representations in different ways~\cite{guo2018countering,xie2018mitigating,raff2019barrage,song2018pixeldefend,xie2019feature}. Feature Squeezing detects adversarial inputs through prediction differences after compression and smoothing~\cite{xu2018squeezing}. Stalder et al.~\cite{stalder2026noise} combine noise and bilateral filters. We adopt these components, learn internal stochastic responses with the backbone frozen, and introduce filtering after adaptation. Randomized smoothing and its robust training target certified smoothed classifiers~\cite{cohen2019certified,salman2019provably}; we evaluate individual random predictions rather than a voting-based certified classifier.

\paragraph{{\spaceskip=\fontdimen2\font\relax Gradient divergence and learned randomness.}}
BaRT reports nearly orthogonal gradients across successive PGD iterations under random transformations~\cite{raff2019barrage}. DRIFT~\cite{guesmi2026drift} learns a stochastic ensemble of residual convolutional input filters before a frozen classifier and explicitly separates gradient responses, providing a direct precedent for disrupting consensus. PNI jointly learns noise scales and network weights~\cite{he2019pni}; Learn2Perturb alternates feature-perturbation and network updates~\cite{jeddi2020learn2perturb}. SAP randomizes and rescales internal activations without retraining~\cite{dhillon2018sap}; RSE trains weights with convolutional noise and averages noisy predictions~\cite{liu2018rse}. DIGN fixes the Gaussian input distribution and original weights, learning internal channel-wise responses with a TRADES-style objective and no explicit gradient-divergence loss. We examine gradients across independent noise draws at a fixed input.

\paragraph{{\spaceskip=\fontdimen2\font\relax Adapting existing classifiers.}}
Denoised Smoothing learns a denoiser before a pretrained classifier for randomized-smoothing certification~\cite{salman2020denoised}; HGD uses high-level representations to guide denoising~\cite{liao2018hgd}. RiFT fine-tunes an existing non-robust-critical module on clean data and interpolates weights to improve generalization while preserving robustness~\cite{zhu2023rift}. DIGN instead freezes all original parameters and BN state and trains new scalar residuals.

Learnable activations include APL, PReLU, and parameterized Swish variants~\cite{agostinelli2015apl,he2015prelu,ramachandran2017swish}. Adapters and LoRA adapt pretrained models with few new parameters~\cite{houlsby2019adapters,hu2021lora}. We study their combination and empirical behavior for stochastic robustness.

\paragraph{{\spaceskip=\fontdimen2\font\relax Full-pipeline evaluation.}}
EOT estimates expected attack gradients under random transformations~\cite{athalye2018eot}. AutoAttack and Square provide complementary attacks~\cite{croce2020autoattack,andriushchenko2020square}, while RobustBench promotes standardized evaluation~\cite{croce2021robustbench}. Work on obfuscated gradients and adaptive evaluation stresses attacking the deployed computation under explicit budgets~\cite{athalye2018obfuscated,carlini2019evaluating,tramer2020adaptive}. Evaluation protocols specify the attack ensembles, EOT budgets, and random-prediction statistics.

\section{DIGN: Learning Divergent Stochastic Responses}
\label{sec:method}
\subsection{Model and threat setting}
Let $f_{\theta,\phi}$ denote a classifier, where $\theta$ contains the pretrained backbone weights and BN state, and $\phi$ contains the new adapter parameters. With a deterministic filter composition $T_\eta$ and noise $\xi\sim\mathcal N(0,I)$, the deployed probability vector is
\begin{equation}
p_\phi(x,\xi)=\operatorname{softmax}\!\left[f_{\theta,\phi}\!\left(T_\eta(x+\sigma\xi)\right)\right].
\label{eq:model}
\end{equation}
Noise is independently sampled on each forward pass. Deployment uses one stochastic prediction; $T_\eta=I$ when filtering is absent. The attacker knows all weights, filter parameters, and the noise distribution. It chooses one RGB perturbation before noise and filtering:
\begin{equation}
\delta\in\mathcal D_\epsilon(x)=\{\delta:\|\delta\|_\infty\leq\epsilon,\ x+\delta\in[0,1]^d\}.
\end{equation}
The attacker cannot prescribe fresh inference noise, but can optimize expected loss using EOT. Thus the same adversarial image must transfer across independent noise realizations.

\subsection{Shared directions and EOT sampling cost}
DIGN examines whether the loss-increasing directions of different noisy observations jointly support one input perturbation. Let $\ell_\xi(x)=-\log p_{\phi,y}(x,\xi)$, $J(x)=\E_\xi\ell_\xi(x)$, and $g_\xi=\nabla_x\ell_\xi(x)$. When differentiation and expectation can be exchanged, the largest first-order loss increase over the norm ball is
\begin{equation}
\max_{\|\delta\|_\infty\leq\epsilon}\E_\xi\langle g_\xi,\delta\rangle
=\epsilon\|\E_\xi g_\xi\|_1.
\label{eq:shared}
\end{equation}
With pixel constraints, this expression is an upper bound. Allowing a different perturbation for every realization instead gives $\epsilon\E_\xi\|g_\xi\|_1$. Their ratio, when defined, lies in $[0,1]$ and describes how much gradient sensitivity a shared perturbation can exploit. Both directions and magnitudes matter.

At a fixed input, assume finite gradient second moments, and let $\mu=\E g_\xi$ and $V=\E\|g_\xi-\mu\|_2^2$. Independent draws satisfy $\E\langle g_\xi,g_{\xi'}\rangle=\|\mu\|_2^2$. When the common component is weak and variation occupies many directions, gradient cosines can concentrate near zero. We report absolute cosines as well as signed ones to distinguish this concentration from signed cancellation.

For $K$ independent EOT draws, $\hat g_K=K^{-1}\sum_k g_{\xi_k}$ satisfies
\begin{equation}
\E\|\hat g_K-\mu\|_2^2=\frac{V}{K}.
\label{eq:eotcost}
\end{equation}
Equation~\eqref{eq:eotcost} characterizes the sampling cost of estimating the expected gradient $\mu=\nabla_xJ(x)$: mean-squared error decreases as $1/K$ with the number of samples. Greater cross-noise gradient variation relative to the common gradient energy requires more samples to achieve the same relative estimation accuracy. This provides a basis for analyzing attack effectiveness across EOT budgets.

\subsection{Robust initialization and noise adaptation}
In the first stage, we train the complete network using standard TRADES or PGD-AT to obtain the backbone $\theta$. Stage two fixes $\theta$, inserts zero-output adapters, and optimizes only $\phi$.
Stochastic adaptation includes an inner EOT attack and an outer adapter update. EOT estimates the stochastic inner objective, while freezing specifies the family of functions available to the outer update.

The restricted family $\{f_{\theta,\phi}\}_\phi$ retains pretrained spatial and channel mixing weights while learning scalar responses. Each adapter changes how one activation enters subsequent fixed feature combinations; it does not learn new spatial or channel mixing operations. This family can change input derivatives while retaining the mixing weights.

The pre-activation placement lets adapters adjust both the conditions under which features pass through ReLU and the local gain of positive responses. These control which responses enter subsequent computation and their strength. The backbone continues to supply learned feature combinations, while adapters learn how to read noisy observations. Zero-output initialization starts this adjustment from the original model function, preserving its predictions before adaptation begins.

The desired prediction criteria also differ. Deterministic robustness seeks correct classification for every admissible perturbation. Single-draw stochastic robustness concerns the error probability of a shared perturbation across the noise distribution, $\sup_{\delta\in\mathcal D_\epsilon(x)}q_\phi(x+\delta)$, with $q_\phi$ defined below. It does not require every noise realization to be correct.

\subsection{Channel-wise adapters and optimization}
For each scalar input $u$ in channel $c$, we replace the original activation by
\begin{equation}
\psi_{\phi,c}(u)=\operatorname{ReLU}\bigl(u+f_{\phi,c}(u)\bigr).
\label{eq:adapter}
\end{equation}
Each $f_{\phi,c}$ is an independent $1\!\to\!8\!\to\!8\!\to\!1$ MLP with SiLU hidden activations. Sharing this scalar function across spatial locations lets noisy activations in one channel be remapped consistently. The final weights and bias are zero-initialized. The 25 activation sites in WRN-28-4 contain 3,600 such channels, increasing the parameter count by approximately 6\% relative to the original backbone. Only these new parameters are optimized in stage two.

At differentiable points, Eq.~\eqref{eq:adapter} has derivative
\begin{equation}
\psi'_{\phi,c}(u)=\mathbf 1_{u+f_{\phi,c}(u)>0}\,[1+f'_{\phi,c}(u)].
\label{eq:adapterderivative}
\end{equation}
Noise can change intermediate inputs, residual derivatives, and activation gates, thereby changing backpropagated input gradients.

Backbone weights and both BN parameters and statistics remain frozen. During adaptation, filtering is absent and $\sigma=0.1$ for the main configuration. The inner attack uses a stop-gradient clean target formed by averaging four noisy probability vectors:
\begin{equation}
\bar p(x)=\operatorname{sg}\!\left[\frac 14\sum_{k=1}^4p_\phi(x,\xi_k)\right].
\end{equation}
We use KL-PGD10/EOT4, an $8/255$ budget, and a $2/255$ step size. Initialization adds Gaussian perturbations of standard deviation $0.001$ and projects to the feasible image set. The complete adapters remain in the attack's forward computation. The adapter update objective is:
\begin{equation}
\begin{aligned}
\mathcal L_\phi={}&-\log p_{\phi,y}(x,\xi_c)\\
&+5D_{\mathrm{KL}}\!\left(\operatorname{sg}[p_\phi(x,\xi_c)]\,\|\,p_\phi(x^{\mathrm{adv}},\xi_a)\right).
\end{aligned}
\end{equation}
This is a TRADES-style adaptation objective with $\beta=5$. Outer updates and formal evaluations use full exact gradients.

\subsection{The stochastic prediction objective}
Let $h_\phi(u)=\arg\max f_{\theta,\phi}(T_\eta(u))$. Single-draw error is $q_\phi(z)=\Pr_\xi[h_\phi(z+\sigma\xi)\ne y]$, the Gaussian probability mass of its error region. Adaptation aims to reduce this mass under the worst shared perturbation, without requiring every noise draw to be correct.

\subsection{Post-adaptation inference filtering}
Filtering in Eq.~\eqref{eq:model} is introduced after adaptation. Gaussian noise is added once, followed by 50 identical $5\times 5$ bilateral-filter passes with reflection padding and no output clipping. The range kernel squares the RGB $L_1$ distance; Appendix~\ref{app:filter} gives the exact expression. Parameters are selected on the fixed WRN-28-4 seed 0 validation set of each dataset and reused across its seeds and widths.

\begin{table*}[t]\centering\normalsize
\caption{Clean and adversarial accuracy across datasets and backbones (\%). D/R denote deterministic/randomized models; + denotes additional training data.}
\label{tab:main}
\setlength{\tabcolsep}{2.6pt}
\begin{tabular}{@{}lllrrrrr@{}}\toprule
\multirow{2}{*}{Dataset} & \multirow{2}{*}{Model} & \multirow{2}{*}{Method} & \multicolumn{1}{c}{\multirow{2}{*}{Clean}} & \multicolumn{3}{c}{$L_\infty$, $\epsilon=8/255$} & \multicolumn{1}{c}{$L_2$}\\
\cmidrule(lr){5-7}\cmidrule(lr){8-8}
 & & & & \multicolumn{1}{c}{PGD20} & \multicolumn{1}{c}{AA} & \multicolumn{1}{c}{CW-PGD100} & \multicolumn{1}{c}{PGD20}\\\midrule
\multirow{5}{*}{CIFAR-10} & \multirow{5}{*}{WRN-28-4} & AT-PGD\textsuperscript{D} & {\fontencoding{T1}\fontfamily{ptm}\selectfont 83.37\,\textpm\,0.80} & {\fontencoding{T1}\fontfamily{ptm}\selectfont 52.90\,\textpm\,0.70} & {\fontencoding{T1}\fontfamily{ptm}\selectfont 48.96\,\textpm\,0.38} & {\fontencoding{T1}\fontfamily{ptm}\selectfont 51.20\,\textpm\,0.25} & {\fontencoding{T1}\fontfamily{ptm}\selectfont 61.19\,\textpm\,0.40}\\
 &  & TRADES\textsuperscript{D} & {\fontencoding{T1}\fontfamily{ptm}\selectfont 82.19\,\textpm\,0.93} & {\fontencoding{T1}\fontfamily{ptm}\selectfont 50.26\,\textpm\,0.42} & {\fontencoding{T1}\fontfamily{ptm}\selectfont 46.99\,\textpm\,0.35} & {\fontencoding{T1}\fontfamily{ptm}\selectfont 48.50\,\textpm\,0.46} & {\fontencoding{T1}\fontfamily{ptm}\selectfont 58.18\,\textpm\,0.77}\\
 &  & Noise+Bilat.\textsuperscript{R,+} & {\fontencoding{T1}\fontfamily{ptm}\selectfont \textbf{85.83\,\textpm\,0.09}} & {\fontencoding{T1}\fontfamily{ptm}\selectfont 61.40\,\textpm\,0.11} & {\fontencoding{T1}\fontfamily{ptm}\selectfont 59.40\,\textpm\,0.36} & {\fontencoding{T1}\fontfamily{ptm}\selectfont 59.48\,\textpm\,0.16} & {\fontencoding{T1}\fontfamily{ptm}\selectfont 80.61\,\textpm\,0.13}\\
 &  & \textbf{DIGN}\textsuperscript{R} & {\fontencoding{T1}\fontfamily{ptm}\selectfont 82.31\,\textpm\,0.43} & {\fontencoding{T1}\fontfamily{ptm}\selectfont \textbf{79.57\,\textpm\,0.19}} & {\fontencoding{T1}\fontfamily{ptm}\selectfont \textbf{76.11\,\textpm\,0.47}} & {\fontencoding{T1}\fontfamily{ptm}\selectfont \textbf{78.26\,\textpm\,0.20}} & {\fontencoding{T1}\fontfamily{ptm}\selectfont \textbf{82.22\,\textpm\,0.40}}\\
 &  & \textbf{DIGN} (PGD-AT)\textsuperscript{R} & {\fontencoding{T1}\fontfamily{ptm}\selectfont 79.99\,\textpm\,1.22} & {\fontencoding{T1}\fontfamily{ptm}\selectfont 75.90\,\textpm\,1.06} & {\fontencoding{T1}\fontfamily{ptm}\selectfont 72.49\,\textpm\,1.50} & {\fontencoding{T1}\fontfamily{ptm}\selectfont 73.97\,\textpm\,1.17} & {\fontencoding{T1}\fontfamily{ptm}\selectfont 79.76\,\textpm\,1.27}\\
\cmidrule(lr){1-8}
\multirow{3}{*}{CIFAR-10} & \multirow{3}{*}{WRN-28-10} & AT-PGD\textsuperscript{D} & {\fontencoding{T1}\fontfamily{ptm}\selectfont 80.46\,\textpm\,1.40} & {\fontencoding{T1}\fontfamily{ptm}\selectfont 47.96\,\textpm\,1.16} & {\fontencoding{T1}\fontfamily{ptm}\selectfont 44.56\,\textpm\,1.14} & {\fontencoding{T1}\fontfamily{ptm}\selectfont 46.79\,\textpm\,1.33} & {\fontencoding{T1}\fontfamily{ptm}\selectfont 57.27\,\textpm\,2.21}\\
 &  & TRADES\textsuperscript{D} & {\fontencoding{T1}\fontfamily{ptm}\selectfont \textbf{82.61\,\textpm\,0.62}} & {\fontencoding{T1}\fontfamily{ptm}\selectfont 51.03\,\textpm\,0.17} & {\fontencoding{T1}\fontfamily{ptm}\selectfont 48.16\,\textpm\,0.19} & {\fontencoding{T1}\fontfamily{ptm}\selectfont 49.66\,\textpm\,0.21} & {\fontencoding{T1}\fontfamily{ptm}\selectfont 58.33\,\textpm\,0.28}\\
 &  & \textbf{DIGN}\textsuperscript{R} & {\fontencoding{T1}\fontfamily{ptm}\selectfont 82.09\,\textpm\,0.97} & {\fontencoding{T1}\fontfamily{ptm}\selectfont \textbf{79.99\,\textpm\,0.66}} & {\fontencoding{T1}\fontfamily{ptm}\selectfont \textbf{75.68\,\textpm\,0.97}} & {\fontencoding{T1}\fontfamily{ptm}\selectfont \textbf{78.83\,\textpm\,0.66}} & {\fontencoding{T1}\fontfamily{ptm}\selectfont \textbf{82.07\,\textpm\,0.96}}\\
\cmidrule(lr){1-8}
\multirow{4}{*}{CIFAR-100} & \multirow{4}{*}{WRN-28-4} & AT-PGD\textsuperscript{D} & {\fontencoding{T1}\fontfamily{ptm}\selectfont 54.09\,\textpm\,0.37} & {\fontencoding{T1}\fontfamily{ptm}\selectfont 28.48\,\textpm\,0.28} & {\fontencoding{T1}\fontfamily{ptm}\selectfont 24.08\,\textpm\,0.15} & {\fontencoding{T1}\fontfamily{ptm}\selectfont 26.20\,\textpm\,0.09} & {\fontencoding{T1}\fontfamily{ptm}\selectfont 36.14\,\textpm\,0.60}\\
 &  & TRADES\textsuperscript{D} & {\fontencoding{T1}\fontfamily{ptm}\selectfont 56.71\,\textpm\,1.63} & {\fontencoding{T1}\fontfamily{ptm}\selectfont 27.77\,\textpm\,0.95} & {\fontencoding{T1}\fontfamily{ptm}\selectfont 23.20\,\textpm\,1.42} & {\fontencoding{T1}\fontfamily{ptm}\selectfont 24.37\,\textpm\,1.33} & {\fontencoding{T1}\fontfamily{ptm}\selectfont 34.62\,\textpm\,0.14}\\
 &  & Noise+Bilat.\textsuperscript{R,+} & {\fontencoding{T1}\fontfamily{ptm}\selectfont \textbf{60.69\,\textpm\,0.56}} & {\fontencoding{T1}\fontfamily{ptm}\selectfont 35.51\,\textpm\,0.34} & {\fontencoding{T1}\fontfamily{ptm}\selectfont 34.11\,\textpm\,0.51} & {\fontencoding{T1}\fontfamily{ptm}\selectfont 33.96\,\textpm\,0.37} & {\fontencoding{T1}\fontfamily{ptm}\selectfont \textbf{53.78\,\textpm\,0.55}}\\
 &  & \textbf{DIGN}\textsuperscript{R} & {\fontencoding{T1}\fontfamily{ptm}\selectfont 52.72\,\textpm\,0.47} & {\fontencoding{T1}\fontfamily{ptm}\selectfont \textbf{50.64\,\textpm\,0.46}} & {\fontencoding{T1}\fontfamily{ptm}\selectfont \textbf{46.16\,\textpm\,0.34}} & {\fontencoding{T1}\fontfamily{ptm}\selectfont \textbf{49.23\,\textpm\,0.72}} & {\fontencoding{T1}\fontfamily{ptm}\selectfont 52.63\,\textpm\,0.46}\\
\cmidrule(lr){1-8}
\multirow{2}{*}{Tiny-ImageNet} & \multirow{2}{*}{WRN-28-4} & TRADES\textsuperscript{D} & {\fontencoding{T1}\fontfamily{ptm}\selectfont \textbf{47.60\,\textpm\,1.68}} & {\fontencoding{T1}\fontfamily{ptm}\selectfont 20.70\,\textpm\,0.62} & {\fontencoding{T1}\fontfamily{ptm}\selectfont 15.43\,\textpm\,0.78} & {\fontencoding{T1}\fontfamily{ptm}\selectfont 16.32\,\textpm\,0.87} & {\fontencoding{T1}\fontfamily{ptm}\selectfont 36.54\,\textpm\,0.74}\\
 &  & \textbf{DIGN}\textsuperscript{R} & {\fontencoding{T1}\fontfamily{ptm}\selectfont 42.30\,\textpm\,1.04} & {\fontencoding{T1}\fontfamily{ptm}\selectfont \textbf{41.20\,\textpm\,1.02}} & {\fontencoding{T1}\fontfamily{ptm}\selectfont \textbf{35.47\,\textpm\,1.11}} & {\fontencoding{T1}\fontfamily{ptm}\selectfont \textbf{40.53\,\textpm\,1.03}} & {\fontencoding{T1}\fontfamily{ptm}\selectfont \textbf{42.32\,\textpm\,0.94}}\\
\bottomrule\end{tabular}
\end{table*}

\begin{table*}[t]\centering\normalsize
\caption{Component and training ablations on CIFAR-10 WRN-28-4 (accuracy in \%).}
\label{tab:ablation}
\setlength{\tabcolsep}{4pt}
\begin{tabular}{@{}lccccc rr@{}}\toprule
Setting & \shortstack{Train\\noise} & \shortstack{Test\\noise} & Adapter & Frozen & Filter & \multicolumn{1}{c}{Clean} & \multicolumn{1}{c}{AA}\\\midrule
TRADES (frozen reference) & \ding{55} & \ding{55} & \ding{55} & \ding{51} & \ding{55} & {\fontencoding{T1}\fontfamily{ptm}\selectfont 82.19\,\textpm\,0.93} & {\fontencoding{T1}\fontfamily{ptm}\selectfont 46.99\,\textpm\,0.35}\\
\midrule
Noise only, no adapter & \ding{55} & \ding{51} & \ding{55} & \ding{51} & \ding{55} & {\fontencoding{T1}\fontfamily{ptm}\selectfont 75.87\,\textpm\,0.32} & {\fontencoding{T1}\fontfamily{ptm}\selectfont 39.04\,\textpm\,0.20}\\
Adapter only, no noise & \ding{55} & \ding{55} & \ding{51} & \ding{51} & \ding{55} & {\fontencoding{T1}\fontfamily{ptm}\selectfont 83.67\,\textpm\,0.68} & {\fontencoding{T1}\fontfamily{ptm}\selectfont 45.45\,\textpm\,0.10}\\
\midrule
Joint from scratch (noise + MLP) & \ding{51} & \ding{51} & \ding{51} & \ding{55} & \ding{55} & {\fontencoding{T1}\fontfamily{ptm}\selectfont \textbf{84.21\,\textpm\,1.01}} & {\fontencoding{T1}\fontfamily{ptm}\selectfont 45.95\,\textpm\,0.32}\\
\midrule
DIGN, noise removed at inference & \ding{51} & \ding{55} & \ding{51} & \ding{51} & \ding{55} & {\fontencoding{T1}\fontfamily{ptm}\selectfont 81.90\,\textpm\,1.37} & {\fontencoding{T1}\fontfamily{ptm}\selectfont 37.38\,\textpm\,2.03}\\
\midrule
DIGN, no inference filtering & \ding{51} & \ding{51} & \ding{51} & \ding{51} & \ding{55} & {\fontencoding{T1}\fontfamily{ptm}\selectfont 82.97\,\textpm\,0.21} & {\fontencoding{T1}\fontfamily{ptm}\selectfont 69.99\,\textpm\,0.83}\\
DIGN (full) & \ding{51} & \ding{51} & \ding{51} & \ding{51} & \ding{51} & {\fontencoding{T1}\fontfamily{ptm}\selectfont 82.31\,\textpm\,0.43} & {\fontencoding{T1}\fontfamily{ptm}\selectfont \textbf{76.11\,\textpm\,0.47}}\\
\bottomrule\end{tabular}
\end{table*}

\begin{figure*}[t]
\centering
\includegraphics[width=\textwidth]{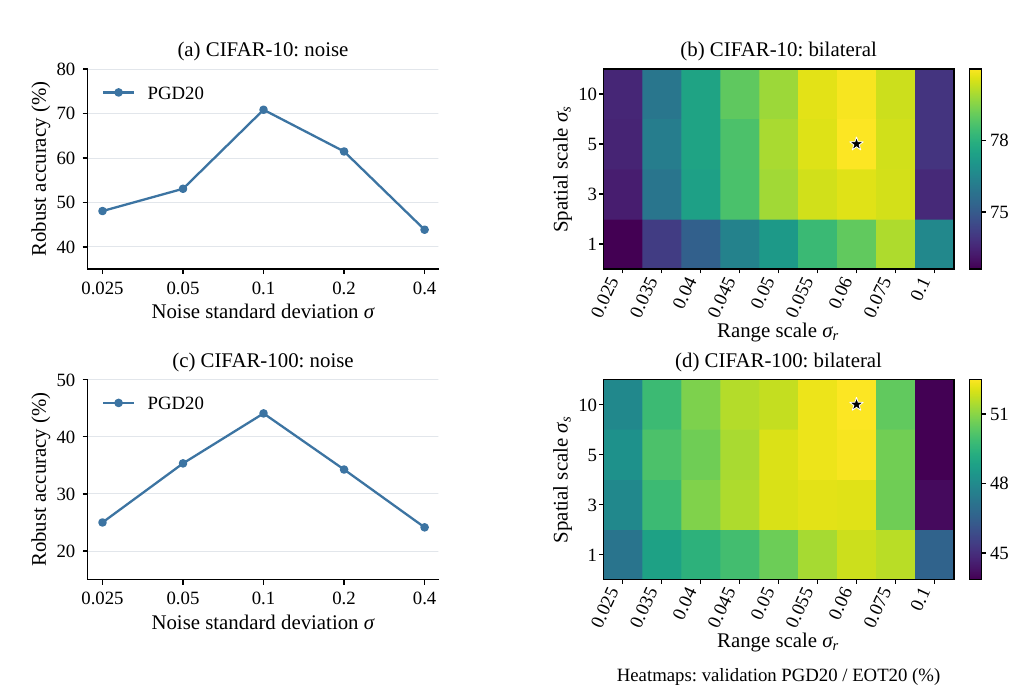}
\caption{Noise-strength sweeps (a,c) and bilateral-filter parameter heatmaps (b,d) on CIFAR-10/100.}
\label{fig:parameter-sensitivity}
\end{figure*}

\section{Experiments}
\subsection{Data, models, and evaluation settings}
We evaluate on CIFAR-10, CIFAR-100~\cite{krizhevsky2009cifar}, and Tiny-ImageNet. We conduct extensive experiments on CIFAR-10/100 and further evaluate transferability to higher-resolution images and a larger label space on Tiny-ImageNet. CIFAR-10 and CIFAR-100 each contain 50,000 training images and 10,000 official test images, covering 10 and 100 classes at $32\times32$ resolution. We use all 50,000 training images and divide the official test set into 1,000 fixed validation images and 9,000 test images. Tiny-ImageNet contains 200 classes at $64\times64$ resolution. We use all 100,000 training images and split its 10,000 labeled official validation images into 1,000 validation images and 9,000 test images. Validation selects checkpoints and filter parameters; final results use the corresponding 9,000-image test split.

We use Wide Residual Networks~\cite{zagoruyko2016wide}: WRN-28-4 on all three datasets and WRN-28-10 additionally on CIFAR-10. All experiments use three random seeds. CIFAR TRADES baselines use $\beta=5$ and 200 epochs of cosine scheduling. The main table reports their validation-best checkpoints, while DIGN adapts each final epoch 200 checkpoint. Checkpoint selection and hyperparameter selection use only the validation set; the test set is reserved for final evaluation. Detailed settings appear in the appendix.

All perturbations act on the original RGB image. The $L_\infty$ budget is $8/255$. PGD20 uses CE loss, step size $2/255$, one uniform random start, and higher-EOT-loss candidate selection across iterations. CW-PGD100 applies 100 PGD updates with truncated CW-margin loss under the same budget and step size~\cite{carlini2017towards,madry2018towards}. $L_2$-PGD20 uses radius $0.5$, step size $0.1$, and one random start inside the ball. Randomized models are attacked with exact gradients through the full pipeline; EOT settings appear in the appendix. After fixing each attack image, we report mean single-prediction accuracy over 16 independent fresh-noise draws.

Randomized AA uses APGD-CE and APGD-DLR, each with 100 steps and one restart. Deterministic AA uses APGD-CE and APGD-T with EOT1. AA reports the library's internal cumulative robust decisions. Clean accuracy is evaluated on clean images. Cells summarize complete 9,000-image results using the mean and sample standard deviation across seeds.

\subsection{Enhancing existing robust models}
Table~\ref{tab:main} summarizes results across datasets, backbones, and training methods. DIGN with WRN-28-4 achieves $79.57\%$ PGD20 and $76.11\%$ AA accuracy on CIFAR-10, and $50.64\%$ and $46.16\%$ on CIFAR-100. Compared with the same-width TRADES entries, AA increases by 29.13 and 22.96 percentage points, respectively. The Noise+Bilat. reproduction uses one million additional synthetic images, as indicated in the table.

The gains also extend to a wider backbone and other attacks. CIFAR-10 WRN-28-10 achieves $79.99\%$ PGD20, $75.68\%$ AA, and $78.83\%$ CW-PGD100 accuracy, with $82.09\%$ clean accuracy. Under $L_2$-PGD20, CIFAR-10 WRN-28-4 and WRN-28-10 achieve $82.22\%$ and $82.07\%$, while CIFAR-100 WRN-28-4 reaches $52.63\%$. These results show a substantially narrower gap between clean and adversarial accuracy.

Both backbone widths benefit substantially, indicating that lightweight adaptation can exploit robust representations already available at different model capacities. On CIFAR-100 in particular, the original models have a large clean--robust gap, which narrows considerably after adaptation. The resulting system preserves more of its recognition performance under attack.

\subsection{From randomization to learned adaptation}
Table~\ref{tab:ablation} examines the components and training configurations. Under the same randomized AA protocol, direct noise achieves $39.04\%$ accuracy. Learning internal adapters raises unfiltered accuracy to $69.99\%$, and post-adaptation filtering further raises it to $76.11\%$. Learning to adapt to random inputs is therefore an important source of the overall gain.

Adversarial joint training from scratch achieves $84.21\%$ clean accuracy and $45.95\%$ AA accuracy, compared with $82.97\%$ and $69.99\%$ for unfiltered DIGN. Their PGD20/EOT20 accuracies are $49.45\%$ and $70.83\%$, respectively. Adapting noisy responses on an existing frozen robust backbone achieves higher adversarial accuracy.

The reversal of clean and robust rankings is particularly informative: joint training recognizes noisy images better, whereas DIGN withstands attacks better. Optimizing noisy recognition and forming an effective defense are therefore distinct objectives. DIGN retains the robust initialization and concentrates its additional degrees of freedom on adapting random observations, consistent with its higher robustness at comparable clean accuracy.

Training adapters without noise yields $45.45\%$ AA accuracy; removing noise from unfiltered DIGN at inference yields $37.38\%$. Together with the complete method, these ablations demonstrate the synergy of noise, adapters, and inference filtering under the proposed design.

\begin{table}[!htbp]
\centering\normalsize
\caption{Noise and bilateral-filter parameters for the three datasets.}
\label{tab:preferred-parameters}
\setlength{\tabcolsep}{4pt}
\begin{tabular}{@{}lcccc@{}}\toprule
Dataset & Resolution & $\sigma$ & $\sigma_r$ & $\sigma_s$\\\midrule
CIFAR-10 & $32\times32$ & 0.1 & 0.06 & 5\\
CIFAR-100 & $32\times32$ & 0.1 & 0.06 & 10\\
Tiny-ImageNet & $64\times64$ & 0.2 & 0.1 & 3\\
\bottomrule
\end{tabular}
\end{table}

\begin{figure}[t]
\centering
\includegraphics[width=\linewidth]{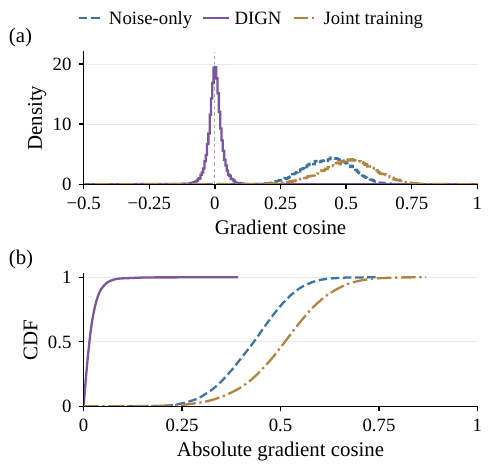}
\caption{Gradient-cosine densities (a) and absolute-cosine cumulative distributions (b) across methods.}
\label{fig:gradients}
\medskip
\begingroup\centering\normalsize
\captionof{table}{Cross-noise gradient-cosine statistics across methods.}
\label{tab:cosine}
\setlength{\tabcolsep}{3pt}
\begin{tabular}{@{}lrrr@{}}\toprule
Statistic & \multicolumn{1}{c}{Noise-only} & \multicolumn{1}{c}{DIGN} & \multicolumn{1}{c}{Joint}\\\midrule
$\E[\cos]$ & 0.429054 & 0.000951 & 0.507081\\
$\E[|\cos|]$ & 0.429054 & 0.019887 & 0.507083\\
$\E[\cos^2]$ & 0.191979 & 0.000796 & 0.268119\\
$|\cos|<0.05$ (\%) & 0.00 & 93.16 & 0.01\\
$|\cos|<0.10$ (\%) & 0.01 & 99.29 & 0.02\\
\bottomrule\end{tabular}
\par\endgroup

\end{figure}

\begin{figure}[t]
\centering
\includegraphics[width=\linewidth]{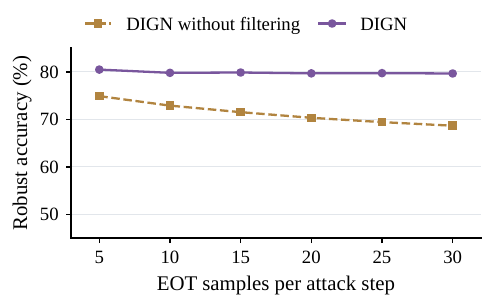}
\caption{PGD20 accuracy versus EOT samples for DIGN with and without filtering.}
\label{fig:eot}
\end{figure}

\subsection{Noise strength and inference enhancement}
The left panels of Fig.~\ref{fig:parameter-sensitivity} show noise-strength sweeps without filtering. Among the tested values, $\sigma=0.1$ achieves the highest three-seed mean PGD20 accuracy on both datasets; smaller and larger noise strengths yield lower accuracy.

The right panels vary filter parameters while holding adapted weights fixed. Each dataset evaluates 44 combinations of 11 range scales and four spatial scales; the heatmaps display 36 combinations. Selection maximizes validation PGD20/EOT20 accuracy, breaking ties by clean accuracy and then smaller parameters. The selected filter is reused across seeds and widths within each dataset. Filtering further improves AA by approximately 6.13 percentage points on CIFAR-10 WRN-28-4 and 3.10 points on CIFAR-100.

The noise-accuracy peaks and high-accuracy heatmap regions in Fig.~\ref{fig:parameter-sensitivity} are similar for CIFAR-10 and CIFAR-100. Together with Table~\ref{tab:preferred-parameters} below the figure, these results show closely matching preferences for the two $32\times32$ datasets: both favor $\sigma=0.1$ and $\sigma_r=0.06$, with similar performance at $\sigma_s=5$ and 10. In contrast, $64\times64$ Tiny-ImageNet uses $\sigma=0.2$, $\sigma_r=0.1$, and $\sigma_s=3$, favoring larger noise and range scales and a smaller spatial scale than the CIFAR datasets.

The intermediate noise optimum suggests a balance between observation diversity and recognition stability: weak noise produces limited input variation, while excessive noise burdens recognition. The continuous high-accuracy heatmap regions also indicate tolerance around effective filter settings. Together, these observations support similar settings for the CIFAR datasets and separate parameter selection for Tiny-ImageNet.

\subsection{Cross-noise divergence and EOT estimation}
Figure~\ref{fig:gradients} presents the unfiltered gradient diagnostic: 100 fixed CIFAR-10 images, 32 independent Gaussian draws per image, and 496 pairs of exact input gradients.

DIGN has mean absolute cosine $0.0199$, compared with $0.4291$ for noise only and $0.5071$ for joint training. Mean squared cosines are $0.000796$, $0.1920$, and $0.2681$, respectively. Both statistics show that DIGN's cross-noise gradients concentrate around nearly orthogonal directions.

Noise-only and joint-training models retain stronger alignment, whereas DIGN exhibits pronounced divergence. Different learning configurations thus yield distinct local attack responses under randomized inputs. Alongside the ablation results, these distributions characterize how learned adaptation changes the model's response to noise.

Equation~\eqref{eq:eotcost} connects this analysis to attack sampling: greater stochastic gradient variation relative to the common gradient energy requires more EOT samples to achieve the same relative estimation accuracy. We next vary the EOT budget to examine its effect on attack performance directly.

\subsection{Stronger EOT and inference filtering}
Figure~\ref{fig:eot} compares the same model and 9,000 test images with and without 50 filter passes. Unfiltered PGD20 accuracy is $74.90\%$, $70.33\%$, and $68.67\%$ at EOT5, 20, and 30, showing that additional samples reveal more effective attack directions.

After bilateral filtering, exact-gradient attack accuracy decreases from $80.49\%$ at EOT5 to $79.67\%$ at EOT30, a reduction of 0.82 percentage points, compared with 6.23 points without filtering. Each point averages 16 fresh-noise single predictions after fixing attacks. The curves differ in both accuracy and sensitivity to the sampling budget: unfiltered accuracy keeps declining, while filtered accuracy levels off after EOT10. Filtering thus provides both higher adversarial accuracy and greater stability as attack sampling increases.

\subsection{Pretraining objectives and Tiny-ImageNet}
DIGN also enhances PGD-AT backbones. Each CIFAR-10 WRN-28-4 seed starts from its final epoch 200 checkpoint. Backbone and BN remain frozen during TRADES-$\beta=5$ adaptation, using $\sigma=0.1$ and the selected CIFAR-10 filter. PGD20, CW-PGD100, and AA reach $75.90\%\!\pm\!1.06\%$, $73.97\%\!\pm\!1.17\%$, and $72.49\%\!\pm\!1.50\%$, respectively.

Together with the two WRN widths, the PGD-AT results show that the same adapter architecture remains effective across robust pretraining objectives. The source initialization affects the final accuracy, while the adaptation and inference pipeline remain shared. The remaining 3.62-point AA gap between TRADES and PGD-AT initializations shows that the source backbone continues to influence final performance.

On Tiny-ImageNet, three-seed DIGN reaches $41.20\%$ PGD20, $35.47\%$ AA, $40.53\%$ CW-PGD100, and $42.32\%$ $L_2$-PGD20 accuracy, with $42.30\%$ clean accuracy. The AA gain over the same-width TRADES baseline is 20.04 percentage points. This extends the method to higher-resolution images and a larger label space.

\subsection{Additional adaptive attacks}
\begin{table}[!htbp]\centering\normalsize
\caption{Adaptive-attack accuracy of DIGN on CIFAR-10 WRN-28-4 (\%).}
\label{tab:adaptive}
\setlength{\tabcolsep}{5pt}
\begin{tabular}{@{}lrr@{}}\toprule
Attack & Budget & Robust (\%)\\\midrule
TA-BPDA (aligned) & 20 steps & {\fontencoding{T1}\fontfamily{ptm}\selectfont 75.81\,\textpm\,0.54}\\
Square / EOT20 & 5000 queries & {\fontencoding{T1}\fontfamily{ptm}\selectfont 75.69\,\textpm\,0.53}\\
\bottomrule\end{tabular}
\end{table}

Table~\ref{tab:adaptive} summarizes TA-BPDA, adapted from the noise-and-filter defense~\cite{stalder2026noise}, and Square. Aligned TA-BPDA uses 20 steps, step size $2/255$, and no random start. The attack forward disables noise and retains filtering; the backward pass uses an identity filter Jacobian and exact adapter gradients. Evaluating the last attack iterate through the complete random pipeline yields $75.81\%\!\pm\!0.54\%$ over three seeds. Square uses EOT20 per score and at most 5000 score queries per image, each averaging 20 random predictions. Accuracy averages 16 fresh-noise single predictions after fixing attacks, yielding $75.69\%\!\pm\!0.53\%$ over three seeds.

The two supplementary attacks yield similar robust accuracies, also close to the AA result. TA-BPDA changes the backward path, while Square searches using model scores. Retaining the gain under both search strategies provides further support for the complete defense.

\subsection{Efficient inference}
\begin{table}[!htbp]
\centering\normalsize
\caption{Single-image inference cost on CIFAR-10 (H100, FP32).}
\label{tab:inference-cost}
\setlength{\tabcolsep}{5pt}
\begin{tabular}{@{}lrr@{}}\toprule
Model & Params. (M) & Latency (ms)\\\midrule
WRN-28-4 & 5.849 & 1.041\\
DIGN & 6.198 & 1.363\\
\bottomrule\end{tabular}
\end{table}

Bilateral-filter acceleration is well studied~\cite{paris2006fast}. We measure inference using standard PyTorch compilation and CUDA Graphs (Table~\ref{tab:inference-cost}).

\section{Discussion}
\paragraph{{\spaceskip=\fontdimen2\font\relax Learning stochastic responses from existing robustness.}}
DIGN separates robust pretraining from noisy adaptation: the first stage supplies reusable features, while the second adjusts responses to random observations. Joint training favors clean recognition; adapting a frozen robust backbone yields stronger robustness under the evaluated attacks.

\paragraph{{\spaceskip=\fontdimen2\font\relax Complementary roles of adaptation and filtering.}}
Adaptation changes cross-noise gradient responses; post-adaptation filtering further improves accuracy and stability across EOT budgets. The filter sweep reuses fixed adapter weights and requires no additional training, connecting learned internal responses with deployment-time input processing.

\section{Conclusion}
We introduce DIGN, a parameter-efficient approach to strengthening existing robust classifiers. By freezing the backbone and batch-normalization state and adversarially optimizing lightweight pre-activation residual adapters under Gaussian noise, DIGN avoids retraining the full network or adding training data, with only approximately 6\% additional parameters for WRN-28-4. Staged training enables input noise, internal adaptation, and inference filtering to work together, outperforming direct noise injection, adapter-only training, and joint training from scratch. We observe nearly orthogonal cross-noise gradient responses after adaptation and examine the further benefit of filtering through full-pipeline EOT attacks. Across CIFAR-10, CIFAR-100, and Tiny-ImageNet, DIGN substantially narrows the clean--robust accuracy gap, while retaining high robustness under the evaluated adaptive and black-box attacks. These results show that reusing robust representations and learning internal responses suited to randomized observations offers an effective alternative to scaling model capacity and training data.

\clearpage
\section*{LLM Usage Statement}
OpenAI Codex (GPT-based models) was used for experimental code development and analysis, and manuscript drafting and revision.

{\small\bibliographystyle{ieeenat_fullname}\bibliography{references-en}}
\clearpage
\appendix
\section{Shared Gradients and EOT Estimation}
\label{app:analysis}
For an integrable random gradient, duality between $L_\infty$ and $L_1$ gives
\begin{equation*}
\max_{\|\delta\|_\infty\leq\epsilon}\langle\E g,\delta\rangle
=\epsilon\|\E g\|_1.
\end{equation*}
A maximizing perturbation is $\epsilon\operatorname{sign}(\E g)$, with arbitrary admissible choices at zero coordinates. Adding valid-pixel constraints makes this value an upper bound. Allowing a different perturbation for each noise realization gives $\epsilon\E\|g\|_1$; Jensen's inequality places their ratio in $[0,1]$ when its denominator is positive.

At a fixed input, assume independent gradient samples with finite second moments, and let $\mu=\E g_\xi$ and $V=\E\|g_\xi-\mu\|_2^2$. Define $v_k=g_{\xi_k}-\mu$. Then $\E v_k=0$ and independence gives $\E\langle v_i,v_j\rangle=0$ for $i\ne j$. Consequently,
\begin{equation}
\begin{aligned}
\E\left\|\frac1K\sum_{k=1}^K g_{\xi_k}-\mu\right\|_2^2
&=\frac1{K^2}\sum_{k=1}^K\E\|v_k\|_2^2\\
&=\frac VK.
\end{aligned}
\label{eq:eot}
\end{equation}
For $\mu\ne0$, the relative mean-squared error is $V/(K\|\mu\|_2^2)$, characterizing the number of EOT samples needed for a given relative accuracy. The calculation applies to fresh independent samples at a fixed input or attack-iteration location. Independence also yields $\E\langle g_\xi,g_{\xi'}\rangle=\|\mu\|_2^2$.

\section{Exact Filter Definition}
\label{app:filter}
Let $\mathcal N(i)$ be a $5\times 5$ neighborhood with reflection padding. One bilateral pass is
\begin{equation}
[B_{\sigma_r,\sigma_s}(u)]_i
=\frac{\sum_{j\in\mathcal N(i)}w_{ij}(u)u_j}{\sum_{j\in\mathcal N(i)}w_{ij}(u)},
\end{equation}
with
\begin{equation*}
w_{ij}(u)=\exp\!\left(-\frac{\|i-j\|_2^2}{2\sigma_s^2}\right)
\exp\!\left(-\frac{\|u_i-u_j\|_1^2}{2\sigma_r^2}\right).
\end{equation*}
For the RGB range term, absolute channel differences are summed before squaring. The final filter is $T_\eta=B_{\sigma_r,\sigma_s}^{50}$. Each pass recomputes weights from the previous pass's output, using the same parameters for all 50 passes. Gaussian noise is added only once before the first pass, not between passes. Outputs are not clipped before or after filtering.

Table~\ref{tab:inference-cost} uses CIFAR-10 inputs on an H100 in FP32, with TF32 and mixed precision disabled. Both models use CUDA Graph replay; DIGN's adapter arithmetic and bilateral passes use PyTorch compilation. DIGN includes one fresh Gaussian draw and 50 filter passes. Latency is the mean of three measurements after warm-up, with inputs resident on the GPU; loading and compilation are excluded.

\section{Training and Evaluation Details}
\label{app:sources}
\label{app:protocol}
Stage one obtains the backbone through standard adversarial training for 200 epochs, using SGD with a cosine learning-rate schedule and an initial rate of $0.1$. TRADES uses $\beta=5$, with standard random cropping and horizontal flipping.

The recommended unified reproduction recipe starts from the final checkpoint at the end of stage one, freezes the backbone and BN, and optimizes only the added adapters. The main adaptation stage has a 200-epoch budget with Adam initialized at $0.01$. A checkpoint is selected by the sum of validation clean and robust accuracies, then refined for 20 further epochs at a learning rate of $0.001$.

Evaluation uses 9,000 test images after excluding the fixed 1,000-image validation set. Stochastic models are attacked through the complete forward pipeline with exact-gradient EOT: EOT20 for CIFAR and EOT10 for Tiny-ImageNet. Randomized models use AutoAttack's AA-rand configuration (\texttt{version='rand'}), comprising APGD-CE and APGD-DLR. PGD and CW report mean single-prediction accuracy over 16 independent fresh-noise draws after fixing the adversarial image; AA reports its internal cumulative robust decisions. Filter parameters are selected on validation data only.

\section{Learned Residual MLP Maps}
\label{app:learned-mlp}
\begin{figure*}[t]
\centering\includegraphics[width=\textwidth]{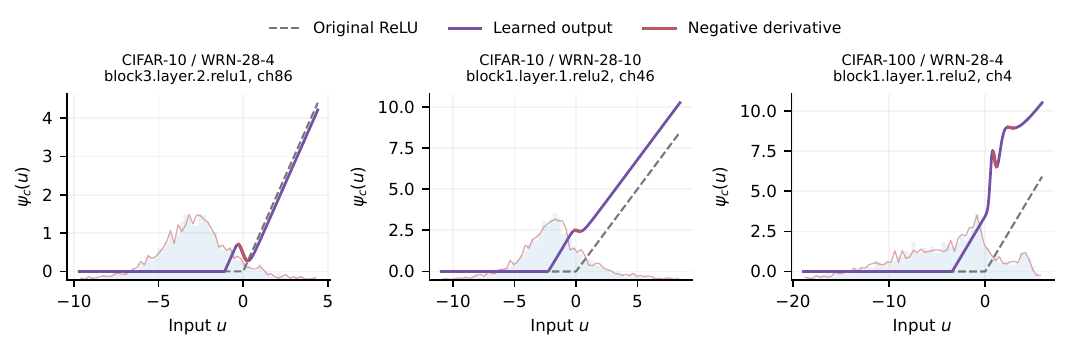}
\caption{Learned scalar activations (purple) versus ReLU (gray dashed); red segments indicate negative derivatives at positive outputs. Columns: CIFAR-10 WRN-28-4, CIFAR-10 WRN-28-10, and CIFAR-100 WRN-28-4, all seed0. Examples have median negative-derivative occupancy among channels exceeding 1\% occupancy. Each panel spans pooled input 0.5--99.5\% quantiles. Blue and red show clean and own-attack input occupancy on a separate auxiliary scale.}
\label{fig:mlp-functions}
\end{figure*}
\begin{table*}[t]
\centering\normalsize
\caption{Channel-map statistics (percent; mean$\pm$sample SD across three seeds). Non-affine error is each model's channel-median relative RMS error of the affine residual fit; gate flips and negative derivatives are measured on sampled inputs.}\label{tab:mlp-functions}
\setlength{\tabcolsep}{6pt}
\begin{tabular*}{\textwidth}{@{\extracolsep{\fill}}lrrr@{}}\toprule
Dataset / width & Non-affine & Gate flips & Neg. inputs\\
\midrule
CIFAR-10 / WRN-28-4 & $22.03\pm0.21$ & $17.76\pm0.43$ & $0.61\pm0.04$\\
CIFAR-10 / WRN-28-10 & $23.25\pm0.35$ & $16.28\pm0.71$ & $0.60\pm0.05$\\
CIFAR-100 / WRN-28-4 & $22.07\pm3.03$ & $19.96\pm0.75$ & $0.75\pm0.06$\\
\bottomrule\end{tabular*}

\end{table*}
\paragraph{Scope and functions.}
We inspect nine main-table adapters: CIFAR-10 WRN-28-4/28-10 and CIFAR-100 WRN-28-4, each with three training seeds. Each channel retains an independent $1\!\to\!8\!\to\!8\!\to\!1$ scalar MLP with SiLU hidden activations, shared across spatial positions. We examine the residual $f_c(u)$ and complete output $\psi_c(u)=\operatorname{ReLU}(u+f_c(u))$.

Each model uses 100 class-balanced test images and its own saved PGD20/EOT20 adversarial images, with four fresh-noise draws and the final 50-pass filter. Eight matched spatial positions per image, channel and draw yield 3,200 local inputs per condition. Clean and adversarial forwards share paired noise; subsequent samples advance the random stream normally. ReLU and adapted outputs are compared at the same local input $u$.

\paragraph{Departure from an affine residual.}
We fit $f_c(u)\approx a_cu+b_c$ on each channel's clean inputs and compute
\begin{equation*}
 e_{f,c}=\frac{\min_{a,b}\operatorname{RMS}_u[f_c(u)-au-b]}
 {\operatorname{RMS}_u[f_c(u)]}.
\end{equation*}
Table~\ref{tab:mlp-functions} takes the channel median for each model, then aggregates three seeds. Median relative fitting errors of 22--23\% quantify the nonlinear component of the learned residuals. The plots show the functions over observed input quantiles.

\paragraph{Gating and response gain.}
Figure~\ref{fig:mlp-functions} shows threshold shifts, input-dependent gains, and local descending segments. Gates change on approximately 16--20\% of identical local inputs. For CIFAR-10 WRN-28-4, on-to-off and off-to-on changes occur on 12.95\% and 4.81\%, respectively. The adapters thus both suppress and activate responses. Where differentiable,
\begin{equation*}
 \psi'_c(u)=\mathbf1_{u+f_c(u)>0}\,[1+f'_c(u)].
\end{equation*}
With convolutional weights fixed, learned gates and local gains change the backpropagated input derivatives.

\paragraph{Negative slopes and occupancy.}
Respectively 11.05\%, 9.33\%, and 13.77\% of channels have negative derivatives on more than 1\% of their sampled inputs. With channels weighted equally, the fractions of inputs occupying negative-slope regions are 0.61\%, 0.60\%, and 0.75\%. The paired input distributions locate these regions and show their occupancy under clean and adversarial inputs.

\paragraph{Interpretation.}
These observations characterize \emph{channelwise nonlinear response and gate remapping}. Each adapter remaps scalar activations within its channel, changing the responses delivered to subsequent fixed feature combinations. The measurements describe the final filtered inference distribution, following adapter training on Gaussian-noisy inputs. Function shapes, gate-change frequencies, and input occupancy provide complementary views of the learned adaptation.

\end{document}